\documentclass{article} 
\usepackage{iclr2026_conference,times}
\usepackage{graphicx}
\usepackage{wrapfig}

\usepackage{booktabs}
\usepackage{array}
\usepackage{siunitx}
\usepackage{caption}
\usepackage{multirow} 
\usepackage{tabularx}
\usepackage{array} 
\usepackage{makecell}
\usepackage{algorithm}
\usepackage{algpseudocode}
\usepackage{amsmath}
\usepackage{subcaption}

\usepackage{amsmath,amsfonts,bm}

\def\eqref#1{equation~\ref{#1}}

\def\1{\bm{1}}

\DeclareMathAlphabet{\mathsfit}{\encodingdefault}{\sfdefault}{m}{sl}
\SetMathAlphabet{\mathsfit}{bold}{\encodingdefault}{\sfdefault}{bx}{n}

\usepackage{hyperref}
\usepackage{url}
\iclrfinaltrue

\title{TimeFlow: Towards Stochastic-Aware and Efficient Time Series Generation via Flow Matching Modeling}

\author{
Panjing He \quad
Mingyue Cheng$^{\dagger}$ \quad
Li Li \quad
XiaoHan Zhang \\
State Key Laboratory of Cognitive Intelligence, University of Science and Technology of China, Hefei, China \\
\texttt{\{hepanjing,lili0516,zxh2519826485\}@mail.ustc.edu.cn, mycheng@ustc.edu.cn}
}

\usepackage{float}
\begin{document}

\maketitle

\begin{abstract}


Generating high-quality time series data has emerged as a critical research topic due to its broad utility in supporting downstream time series mining tasks. A major challenge lies in modeling the intrinsic stochasticity of temporal dynamics, as real-world sequences often exhibit random fluctuations and localized variations. 
While diffusion models have achieved remarkable success, their generation process is computationally inefficient, often requiring hundreds to thousands of expensive function evaluations per sample. 
Flow matching has emerged as a more efficient paradigm, yet its conventional ordinary differential equation (ODE)-based formulation fails to explicitly capture stochasticity, thereby limiting the fidelity of generated sequences. By contrast, stochastic differential equation (SDE) are naturally suited for modeling randomness and uncertainty. Motivated by these insights, we propose \textbf{TimeFlow}, a novel SDE-based flow matching framework that integrates a encoder-only architecture. 
Specifically, we design a component-wise decomposed velocity field to capture the multi-faceted structure of time series and augment the vanilla flow-matching optimization with an additional stochastic term to enhance representational expressiveness. TimeFlow is flexible and general, supporting both unconditional and conditional generation tasks within a unified framework. Extensive experiments across diverse datasets demonstrate that our model consistently outperforms strong baselines in generation quality, diversity, and efficiency. 
The code is available at \url{https://github.com/PanJingHe/TimeFlow}.


\end{abstract}

\section{Introduction}

Time series generation plays an important role across many real-world domains, including finance, healthcare, and energy management~\citep{intro1,finance,healthcare,energy}, where reliable modeling and simulation are essential for decision making, intelligent management, and risk assessment~\citep{survey}. Consequently, time series generation has attracted considerable research attention. Unlike images or text, time series are characterized by long-range dependencies~\citep{long_dependency} and inherent stochasticity~\citep{time_r1,cheng2025convtimenet}, which make their generative modeling particularly challenging. A central challenge in this field lies in capturing the inherent stochasticity of temporal dynamics, which arises from noise and perturbations and is crucial for reproducing realistic variability~\citep{formertime,cheng2025instructime}.

\begin{figure}[t]
    \centering
    \includegraphics[width=\textwidth]{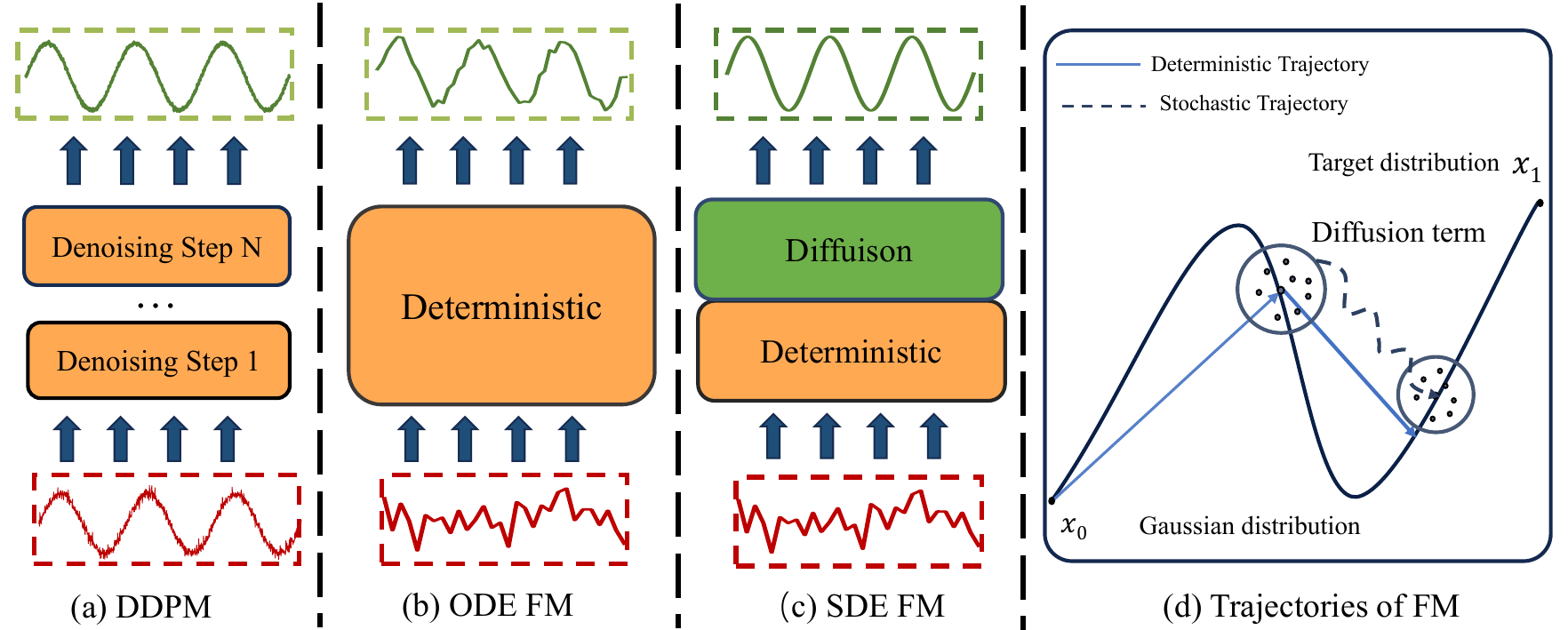}
    \caption{Comparison of generative paradigms. (a) DDPM rely on iterative denoising. (b) ODE-based flow matching produces deterministic trajectories. (c) SDE-based flow matching yields stochastic trajectories that capture uncertainty via diffusion terms. (d) Illustration of deterministic versus stochastic trajectories under flow matching.}
    \label{main_figure1}
\end{figure}

A variety of generative frameworks have been explored for time series modeling, including generative adversarial networks (GANs)~\citep{gan} and variational autoencoders (VAEs)~\citep{intro5,tao2025values}, which provide early probabilistic and adversarial approaches for synthesizing temporal data. 
More recently, denoising diffusion probabilistic models (DDPMs)~\citep{intro3,intro4} have achieved remarkable success in the broader field of generative modeling, powering breakthroughs in image, speech, and scientific data synthesis~\citep{ddpm_image,ddpm_speech,diffusion-ts,timereasoner}. They have also demonstrated strong potential for time series tasks, setting new baselines for generative quality. Despite their impressive flexibility, DDPMs are computationally inefficient, as iterative denoising requires hundreds to thousands of reverse steps (Figure~\ref{main_figure1}(a)). This inefficiency is particularly problematic in time series generation, where longer sequences further amplify the computational burden, limiting the practicality of DDPMs in large-scale or latency-sensitive scenarios. To overcome this efficiency bottleneck, flow matching (FM)~\citep{flowmatching} has emerged as a promising alternative by directly parameterizing continuous generative trajectories through velocity fields, offering both stable training objectives and efficient sampling mechanisms.

However, most existing FM applications adopt an ordinary differential equation (ODE) formulation, where trajectories are determined by deterministic velocity fields, as illustrated in Figure~\ref{main_figure1}(b). While effective in modeling global structure, this ODE-based FM model suffers from two key limitations for time series generation. First, it does not explicitly model stochasticity, resulting in synthetic sequences that lack variability and fail to reproduce rare but important fluctuations. Second, by constraining the generative dynamics to a deterministic flow, the model cannot adequately represent predictive uncertainty or adapt to heterogeneous temporal regimes. In other words, although FM improves efficiency, the ODE formulation sacrifices the ability to faithfully reflect the stochastic nature of real-world temporal processes. These shortcomings motivate extending FM to a stochastic differential equation (SDE) formulation, where diffusion terms (Figure~\ref{main_figure1}(c)) enable richer variability, uncertainty-aware trajectories, and improved robustness to random perturbations.

To address these challenges, we propose TimeFlow, an SDE-based flow matching framework for efficient and stochasticity aware time series generation (Figure~\ref{main_figure1}(d)). By incorporating a diffusion term into the generative process, TimeFlow explicitly captures randomness and produces higher-quality trajectories with faithful uncertainty modeling. To further preserve temporal structures, we employ a transformer-based encoder with decomposition mechanism, which maintains consistency with underlying dynamics while keeping the framework  and efficient. Moreover, TimeFlow naturally extends to both unconditional and conditional generation tasks, including forecasting and imputation, thereby broadening its applicability in practice.

In summary, our major contributions are as follows:
\begin{itemize}
    \item We propose TimeFlow, a novel time series generation framework that extends flow matching to a stochastic differential equation formulation. By introducing the Stochastic Flow Matching loss, our method explicitly models temporal stochasticity, enabling uncertainty aware generative processes that better capture complex random fluctuations.  

    \item We leverage the flow matching paradigm to fundamentally alleviate the efficiency bottlenecks of diffusion-based approaches, enabling significantly faster generation of high dimensional time series while maintaining both reliability and fidelity.   

    \item We conduct evaluations on both unconditional and conditional generation tasks, including forecasting and imputation, across diverse real world datasets. Experimental results demonstrate that TimeFlow outperforms strong baselines in both quality and efficiency.
\end{itemize}

\section{Related Work}

\paragraph{Time Series Generation.}

Generative models have recently shown strong performance in time series modeling. Early efforts focused on GAN-based methods, such as TimeGAN~\citep{timegan}, which introduces a supervised embedding to preserve temporal dynamics. COT-GAN~\citep{cotgan} improves the stability of training through causal optimal transport and entropic regularization, while GT-GAN~\citep{gtgan} offers a general framework for regular and irregular time series.
VAE-based approaches were later explored. TimeVAE~\citep{timevae} proposes an interpretable structure tailored to time series, and CR-VAE~\citep{crvae} encodes causal dependencies via a sparse critical matrix. More recently, diffusion models have emerged as a leading paradigm. DiffTime~\citep{difftime} decouples constraint specification from training, enabling flexible inference-time adaptation. SSSD~\citep{sssd} and CSDI~\citep{csdi} extend the framework to conditional tasks using self-supervised masking. TimeGrad~\citep{timegrad} applies RNN-guided autoregression, while TSDiff~\citep{tsdiff} enables unconditional generation via self-guidance. Finally, Diffusion-TS~\citep{diffusion-ts} introduces an interpretable design that disentangles temporal semantics.

\vspace{-5pt}
\paragraph{Flow Matching.}
As a promising alternative to diffusion processes, flow matching~\citep{flowmatching} has shown significant potential in generative modeling due to its training stability, flexible trajectory design, and computational efficiency. Flow matching was initially proposed for image generation tasks, where it effectively modeled complex data distributions~\citep{image1,image2,image3,meanflow}. It and its variants, such as rectified flow~\citep{rectifiedflow} and OT flow matching~\citep{otflow}—have since been extended beyond image generation to a variety of domains, including video~\citep{video1,video2,video3}, text~\citep{text1}, and audio~\citep{audio1}. Recently, flow matching has been applied to time series. TFM~\citep{tfm} uses neural SDEs to tackle stochastic and irregularly sampled clinical time series forecasting. SGFM~\citep{sgfm} combines state space models and GNNs for refined anomaly detection. CGFM~\citep{cgfm} models prediction time series forcasting errors by conditional guidance. CFM-TS~\citep{cfm-ts} applies conditional probability paths to neural ODEs for time series modeling.
\section{Preliminaries}

\subsection{Problem Statement}
Let $\mathbf{X}_{1:t} = (x_1, \ldots, x_t) \in \mathbb{R}^{\tau \times d}$ denote an observed time series, where $t$ represents the number of time steps, and $d$ denotes the dimension of each observation. Given the time series dataset $\mathcal{S} = \{ \mathbf{X}_{1:t}^i \}_{i=1}^N$ consisting of $N$ time series samples, the objective of the unconditional generation task is to learn a flow-based generator $G$ to synthesize sequences $\hat{\mathbf{X}}_{1:t}^i = G(\mathbf{S}_i)$, which closely approximates the realistic time series data $\mathbf{X}_{1:t}^i$. And the goal of conditional generation is to generate samples from a conditional distribution $p(\cdot \mid y)$, where $y$ is a control variable that can be any real-world signal and dictates the synthesis.

\subsection{Flow Matching}
 
Flow matching~\citep{flowmatching} is a family of generative models that learn to match probability flows represented by velocity fields between a simple prior distribution and the data distribution. Formally, given data $x_1 \sim p_{\text{data}}(x)$ and a prior noise variable $x_0 \sim p_{\text{prior}}(\epsilon)$, a flow path can be defined as $x_{t} = a_{t}x_1 + b_{t}x_0$, where $a_{t}$ and $b_{t}$ are predefined schedules. The corresponding velocity is given by $v_{t} = \dot{z}_{t} = a_{t}'x + b_{t}'\epsilon$, with $(\cdot)'$ denoting the derivative with respect to time. This is referred to as the conditional velocity, written as $v_{t} = v_{t}(x_{t}\mid x_1)$. A common choice of schedules is $a_{t}=t$ and $b_{t}=1-t$, which yields $v_{t} = x_1 - x_0$.  

Since a given $x_{t}$ and its associated velocity $v_{t}$ can arise from different pairs of $x$ and $\epsilon$, flow mtching essentially models the expectation over all possibilities, which is called the marginal velocity: $\nu(x_{t}, t) = \mathbb{E}_{p(v_{t} \mid x_{t})}[v_{t}].$ A neural network $v_{\theta}$ parameterized by $\theta$ is trained to approximate this marginal velocity field. The conditional flow matching loss is written as:  
\begin{equation}
\mathcal{L}_{\text{CFM}}(\theta) = \mathbb{E}_{t,x_1,x_0} \big[ \| v_{\theta}(x_{t}, t) - v_{t}] \|^{2} \big],
\end{equation}  
where the target $v_{t}$ corresponds to the conditional velocity. By minimizing this objective, the learned velocity field aligns with the expected flow, enabling efficient training of generative models.  

\subsection{Neural Stochastic Differential Equation}
Neural stochastic differential equations describe the evolution of a latent state with both deterministic and stochastic components, given by:
\begin{equation}
dz_t = f_\theta(x_t, t)\,dt + g_\theta(x_t, t)\,dW_t,
\end{equation}
where $f_\theta$ denotes the drift term parameterized by a neural network, $g_\theta$ is the diffusion coefficient, and $W_t$ is a standard Wiener process. The first term $f_\theta(x_t, t)\,dt$ corresponds to the deterministic drift, while the second term $g_\theta(x_t, t)\,dW_t$ represents stochastic perturbations from Brownian motion.

\begin{figure}[t]
    \centering
    \includegraphics[width=\textwidth]{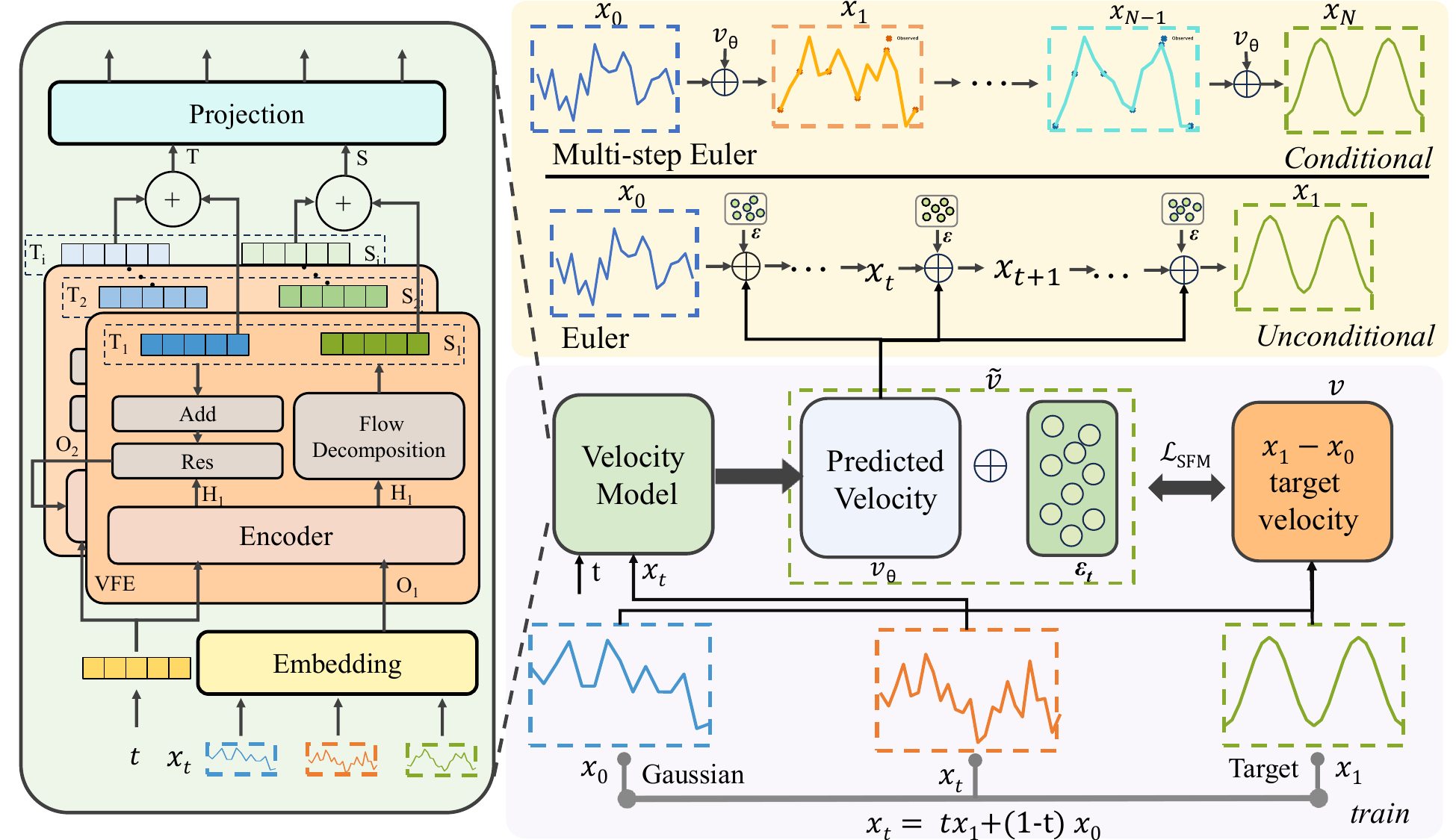}
    \caption{Overview of the proposed TimeFlow architecture }
    \label{main_figure2}
\end{figure}

\section{Methods}

\subsection{Overview}
Time series data often exhibit inherent stochasticity caused by noise and random perturbations. To address this issue, we propose a flow matching framework that explicitly learns velocity fields to align the dynamics between prior noise and target sequences. As illustrated in Figure~\ref{main_figure2}, our method consists of four main components. First, the velocity model, composed of multiple velocity field encoder (VFE) layers, parameterizes the continuous transformation from a Gaussian prior $x_0$ to the real data distribution $x_1$. Second, stochastic optimization objective incorporates perturbations $\epsilon_t$ to capture the inherent randomness of time series. Third, flow decomposition (FD) employs decomposition to disentangle temporal components. Finally, stochastic sampling enables versatile generation.Unconditional generation directly synthesizes sequences from noise, while conditional generation exploits observed data to produce consistent and plausible completions.

\subsection{Velocity Model}

We propose a velocity model consisting of multiple Velocity Field Encoder (VFE) layers, each instantiated as a transformer encoder architecture that captures long dependency and global contextual interactions. At the input stage, a convolutional embedding layer extracts local temporal patterns and projects raw sequences into a high-dimensional representation space. Then, the encoder layer refines temporal representations and applies a decomposition mechanism to disentangle the output into two complementary components. The seasonal component captures high-frequency stochastic dynamics, while the trend component represents low-frequency smooth evolution as a stable baseline. This explicit separation prevents heterogeneous sources of uncertainty from being conflated into a single representation, thereby enabling a more structured parameterization of the velocity field and enhancing robustness to stochastic perturbations.

\paragraph{Flow Decomposition.} 
In our velocity field encoder, the encoder output $H_i$ is first decomposed via moving average into a trend component $t_i$ and a seasonal component $s_i$ for subsequent processing:
\vspace{-2pt}
\begin{equation}
    t_i = \text{AvgPool}(\text{Padding}(H_i)), \qquad
    s_i = H_i - t_i,
\end{equation}
\vspace{-2pt}Given that seasonal patterns are vulnerable to contextual disturbances such as phase shifts and amplitude variations, we refine $s_i$ using a cross-attention (CA) mechanism with the original representation $H_i$, and apply a residual connection to promote stable learning. In comparison, the trend component $t_i$ is inherently more stable; nevertheless, we further modulate it through a multi-scale gated convolutional module that effectively incorporates seasonal information:
\vspace*{-2pt}
\begin{equation}
    S_i = \text{CA}(s_i, H_i) + s_i, \qquad
    T_i = \Big(\sum_{k=1}^{K} \text{Conv1D}_k(S_i)\Big) \odot t_i,
\end{equation}
\vspace*{-2pt}Here, $S_i$ represents the refined seasonal representation, and $T_i$ the modulated trend representation. Subsequently, the residual representation for the next layer is computed by removing both components from the encoder output, whereas the velocity field is generated from the aggregated seasonal–trend terms accumulated over $D$ layers of the encoder:
\vspace*{-2pt}
\begin{equation}
    O_{i+1} = H_i - S_i - T_i, \qquad
    v_{\theta}(x_t, t) = \text{Projection}\Big(\sum_{i=1}^{D} S_i + \sum_{i=1}^{D} T_i\Big),
\end{equation}
where $O_{i+1}$ is the residual representation passed to the $(i{+}1)$-th encoder layer, and $v_{\theta}(x_t, t)$ is the velocity field parameterized by $\theta$ at time step $t$.

\subsection{Stochastic Optimization Objective}

To explicitly account for stochastic perturbations in time series, we introduce the stochastic flow matching (SFM) loss. 
As illustrated in Figure~\ref{main_figure2}, given an intermediate state $x_t$, the velocity model predicts a velocity field $v_{\theta}(x_{t}, t)$. 
To model uncertainty, a Gaussian noise term $\epsilon_t \sim \mathcal{N}(0, \sigma_t^2 I)$ is injected into the prediction, yielding a perturbed velocity representation: $\tilde{v}(x_{t}, t) = v_{\theta}(x_{t}, t) + \epsilon_t.$ The SFM loss is then defined as:
\begin{equation}
    \mathcal{L}_{\text{SFM}} = \mathbb{E}_{t, x_t}\!\left[\left\| \tilde{v}(x_{t}, t) - v_t \right\|^2\right],
\end{equation}
where $v_{t} = x_1 - x_0$ denotes the ground-truth velocity between the initial and terminal states. 
By minimizing $\mathcal{L}_{\text{SFM}}$, the model is encouraged to learn a velocity field that captures not only deterministic dynamics but also stochastic variations, thereby facilitating robust and faithful modeling of uncertain time series.

\subsection{Stochastic Sampling}
As illustrated in Figure~\ref{main_figure2}, we design two distinct sampling flows for time series generation: unconditional and conditional. Both flows rely on the learned velocity field but differ in how stochastic perturbations and observational information are incorporated, thereby providing complementary perspectives on how the model handles uncertainty. In particular, unconditional flow emphasizes the role of randomness in generating diverse trajectories, while conditional flow demonstrates how partial observations can guide the generation process toward consistency with known information.  
\vspace{-5pt}
\paragraph{Unconditional Generation.} 
The unconditional generation process begins from a Gaussian prior $X_0 \sim \mathcal{N}(0, I)$ and evolves according to a stochastic integral equation of the form:
\begin{equation}\setlength{\abovedisplayskip}{0pt}\setlength{\belowdisplayskip}{0pt}
    X_1 =X_0 + \int_{0}^{1} \Big( v_{\theta}(t, X_t)\, dt + \sigma \, dW_t \Big),
\end{equation}
where $v_{\theta}(t, X_t)$ denotes the learned velocity field, and $\sigma \, dW_t$ is a stochastic term driven by Brownian motion that introduces random perturbations. This formulation explicitly establishes a connection between the generative process and stochastic dynamics, ensuring that the synthesized trajectories not only capture the overall structure of time series but also reproduce their intrinsic variability. By integrating both deterministic velocity guidance and stochastic diffusion.  
\vspace{-5pt}
\paragraph{Conditional Generation.}  
The conditional generation process incorporates partial observations of the target sequence as guidance. At each iteration, the integration time $t \in [0,1]$ is reparameterized using a power-based sampling scheme, which controls the pace of noise injection and enables both early stochastic exploration and stable refinement in later stages. The latent state is obtained by interpolating between Gaussian noise and the observed sequence, with entries specified by a partial mask replaced by noisy versions of the ground truth to enforce conditional fidelity. The evolution of the trajectory is then governed by an Euler update rule of the form:
\begin{equation}
    x_{t+1} = x_t + (1 - t)\, v_\theta(x_t, t),
\end{equation}
where $v_\theta(\cdot)$ denotes the learned velocity field. To ensure stability, the outputs are further bounded through clamping. This formulation generates trajectories that remain faithful to observed data while preserving realistic variability in unobserved regions.

\vspace{-5pt}
\section{Experiments}

\subsection{Experimental Setups}

\paragraph{Dataset.} We evaluate our approach on four real-world datasets and two synthetic datasets. The real-world datasets include Stocks, which contains Google stock prices from 2004 to 2019; ETTh, which records 15-minute electricity transformer load and oil temperature measurements; and UCI Energy, which provides household appliance energy consumption data. In addition, the fMRI dataset offers simulated BOLD signals. For synthetic benchmarks, we adopt the Sines dataset, comprising five-dimensional sinusoidal sequences, and the MuJoCo dataset, which generates multivariate time series from a physics-based simulation environment.

\paragraph{Metrics.} We evaluate the quality of the synthesized data using four widely adopted metrics. (1) The discriminative score~\citep{timegan} measures the similarity between real and synthetic data based on a supervised classification task. (2) The predictive score~\citep{timegan} assesses the utility of synthetic data by training a sequence model on synthetic samples and evaluating it on real data. (3) The Context-Fr\'echet Inception Distance (Context-FID)~\citep{psagan} quantifies the sample quality by comparing their contextual representations with those of real data. (4) The correlational score~\citep{psagan} evaluates temporal dependency by computing the absolute error between correlation structures of the real and synthetic data.

\paragraph{Baselines.} For unconditional generation, we compare our model with six baselines, including Diffusion-TS~\citep{diffusion-ts}, TimeGAN~\citep{timegan}, TimeVAE~\citep{timevae}, DiffWave~\citep{diffwave}, DiffTime~\citep{difftime}, and Cot-GAN~\citep{cotgan}. For conditional generation, we compare our model with Diffusion-TS~\citep{diffusion-ts} and CSDI~\citep{csdi}.

\subsection{Unconditional Time Series Generation}

In Table~\ref{tab:results}, we present results for 24-length time series generation, a setting widely adopted in prior studies. TimeFlow consistently outperforms baseline methods across six datasets and nearly all evaluation metrics, with average improvements of 59.9\% in Context-FID and 57.5\% in Discriminative Score over the strongest competitor. We further evaluate TimeFlow-ODE, an ODE-based variant, which achieves competitive performance and attains state-of-the-art results in more than half of the cases. Nonetheless, TimeFlow consistently surpasses TimeFlow-ODE. These findings show that explicitly modeling stochasticity improves robustness and fidelity, highlighting the necessity of capturing intrinsic randomness in time series.
\looseness=-1

\begin{table}[ttt]
  \renewcommand{\arraystretch}{1.1}
  \setlength{\tabcolsep}{5pt} 
  \centering
  \scriptsize
  \caption{Results on multiple time-series datasets (Bold indicates best performance).}
  \begin{tabular}{c|c|c|c|c|c|c|c}
    \toprule
    Metric (↓) & Methods & Sincs & Stocks & ETTh & MuJoCo & Energy & fMRI \\
    \midrule
    \multirow{8}{*}{\makecell{Context-FID \\ Score \\ Lower Better}} 
        & TimeFlow & \textbf{0.002±0.000} & \textbf{0.011±0.004} & \textbf{0.016±0.001} & \textbf{0.012±0.001} & \textbf{0.027±0.003} & \underline{0.122±0.009} \\
        & TimeFlow-ODE & 0.007±0.001 & \underline{0.016±0.004} & \underline{0.082±0.015} & 0.037±0.005 & \underline{0.034±0.003} & 0.170±0.008 \\
        & Diffusion-TS & \underline{0.006±0.000} & 0.147±0.025 & 0.116±0.010 & \underline{0.013±0.001} & 0.089±0.024 & \textbf{0.105±0.006} \\
        & TimeGAN & 0.101±0.014 & 0.103±0.013 & 0.300±0.013 & 0.563±0.052 & 0.767±0.103 & 1.292±0.218 \\
        & TimeVAE & 0.307±0.060 & 0.215±0.035 & 0.805±1.186 & 0.251±0.015 & 1.631±1.142 & 14.449±9.969 \\
        & Diffwave & 0.014±0.002 & 0.232±0.032 & 0.873±0.061 & 0.393±0.041 & 1.031±1.131 & 0.244±0.018 \\
        & DiffTime & 0.006±0.001 & 0.236±0.074 & 0.299±0.044 & 0.188±0.028 & 0.279±0.045 & 0.340±0.015 \\
        & Cot-GAN & 1.337±0.068 & 0.408±0.086 & 0.980±0.071 & 1.094±0.079 & 1.039±0.028 & 7.813±3.550 \\
    \midrule
    \multirow{8}{*}{\makecell{Correlational \\ Score \\ Lower Better}}
        & TimeFlow  & \textbf{0.011±0.001} & \textbf{0.003±0.002} & \textbf{0.027±0.010} & \textbf{0.163±0.032} & \underline{0.576±0.103} & \textbf{0.837±0.010} \\
        & TimeFlow-ODE & 0.017±0.007 & 0.008±0.005 & \underline{0.048±0.014} & \underline{0.170±0.022} & \textbf{0.573±0.094} & \underline{0.930±0.016} \\
        & Diffusion-TS & \underline{0.015±0.004} & \underline{0.004±0.001} & 0.049±0.008 & 0.193±0.027 & 0.856±1.147 & 1.411±0.42 \\
        & TimeGAN & 0.045±0.010 & 0.063±0.005 & 0.210±0.006 & 0.886±0.039 & 4.010±1.104 & 23.506±2.039 \\
        & TimeVAE & 0.131±0.010 & 0.095±0.008 & 0.111±0.200 & 0.388±0.041 & 1.688±2.226 & 17.292±3.526 \\
        & Diffwave & 0.022±0.005 & 0.030±0.020 & 0.175±0.006 & 0.579±0.018 & 5.001±1.154 & 3.927±0.409 \\
        & DiffTime & 0.017±0.004 & 0.008±0.002 & 0.067±0.005 & 0.128±0.031 & 1.158±0.095 & 1.501±0.048 \\
        & Cot-GAN & 0.049±0.010 & 0.007±0.004 & 0.249±0.009 & 1.041±2.007 & 3.164±1.061 & 26.824±4.449 \\
    \midrule
    \multirow{8}{*}{\makecell{Discriminative \\ Score \\ Lower Better}}
        & TimeFlow  & \textbf{0.004±0.004} & \textbf{0.011±0.010} & \textbf{0.010±0.006} & \textbf{0.005±0.007} & \underline{0.061±0.007} & \textbf{0.101±0.015} \\
        & TimeFlow-ODE & 0.014±0.007 & \underline{0.034±0.018} & \underline{0.023±0.010} & 0.023±0.026 & \textbf{0.060±0.021} & \underline{0.141±0.019} \\
        & Diffusion-TS & \underline{0.006±0.007} & 0.067±0.015 & 0.061±0.009 & \underline{0.008±0.002} & 0.122±0.003 & 0.167±0.023 \\
        & TimeGAN & 0.011±0.008 & 0.102±0.021 & 0.114±0.055 & 0.238±0.068 & 0.236±0.012 & 0.484±0.042 \\
        & TimeVAE & 0.041±0.044 & 0.145±1.120 & 0.209±0.058 & 0.230±1.102 & 0.499±1.000 & 0.476±0.044 \\
        & Diffwave & 0.017±0.008 & 0.232±0.061 & 0.190±0.008 & 0.203±0.096 & 0.493±0.004 & 0.402±0.029 \\
        & DiffTime & 0.013±0.006 & 0.097±0.016 & 0.100±0.007 & 0.154±0.045 & 0.445±0.004 & 0.245±0.051 \\
        & Cot-GAN & 0.254±1.137 & 0.230±0.016 & 0.325±0.099 & 0.426±0.022 & 0.498±0.002 & 0.492±0.018 \\

    \midrule
    \multirow{9}{*}{\makecell{Predictive \\ Score \\ Lower Better}}
        & TimeFlow  & \textbf{0.093±0.000} & \underline{0.037±0.000} & \textbf{0.119±0.002} & \underline{0.008±0.000} & \textbf{0.250±0.000} & \textbf{0.099±0.000} \\
        & TimeFlow-ODE & \underline{0.093±0.000} & \underline{0.037±0.000} & 0.123±0.009 & 0.011±0.000 & \underline{0.250±0.000} & \underline{0.099±0.000} \\
        & Diffusion-TS & \underline{0.093±0.000} & \textbf{0.036±0.000} & \underline{0.119±0.002} & \textbf{0.007±0.003} & \underline{0.250±0.000} & \underline{0.099±0.000} \\
        & TimeGAN & 0.093±0.019 & 0.038±0.001 & 0.124±0.001 & 0.025±1.000 & 0.273±0.004 & 0.126±1.002 \\
        & TimeVAE & \underline{0.093±0.000} & 0.039±0.000 & 0.126±0.004 & 0.012±0.002 & 0.292±0.000 & 0.113±0.003 \\
        & Diffwave & \underline{0.093±0.000} & 0.047±0.001 & 0.130±0.001 & 0.013±0.001 & 0.251±0.000 & 0.101±0.000 \\
        & DiffTime & \underline{0.093±0.000} & 0.038±0.000 & 0.121±0.004 & 0.010±0.000 & 0.252±0.000 & 0.100±0.000 \\
        & Cot-GAN & 0.100±0.000 & 0.047±0.001 & 0.129±0.000 & 0.068±0.009 & 0.259±0.000 & 0.185±0.003 \\
         \cmidrule(){2-8}
        & Original & 0.094±.001 & 0.036±.001 & 0.121±.005 & 0.007±.001 & 0.250±.003 & 0.090±.001 \\
    \bottomrule
  \end{tabular}
  \label{tab:results}
\end{table}

\begin{figure}[h]
    \centering
    \includegraphics[width=\textwidth]{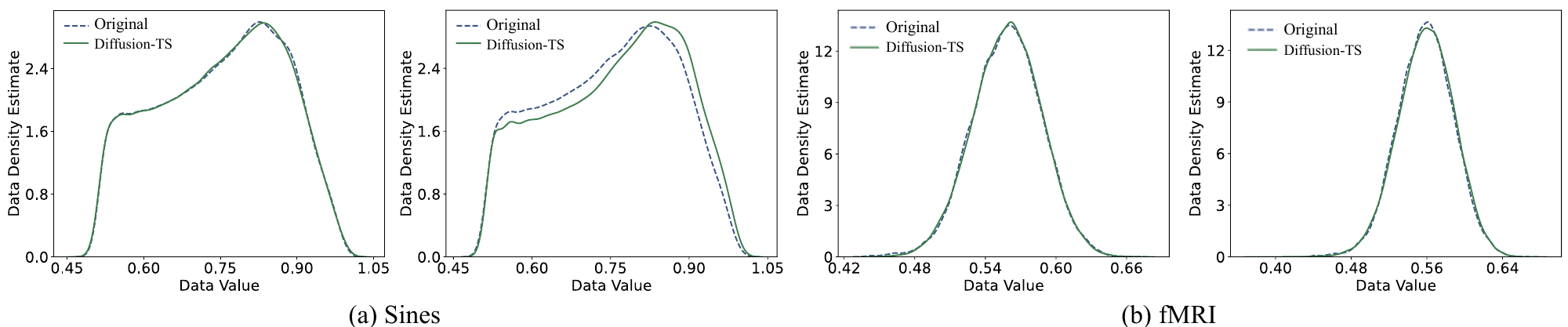}
    \caption{KDE analysis of time series synthesized by TimeFlow and Diffusion-TS}
    \label{tsne_kde}
\end{figure}

\begin{wrapfigure}{r}{0.5\columnwidth} 
    \centering
    \includegraphics[width=0.48\columnwidth]{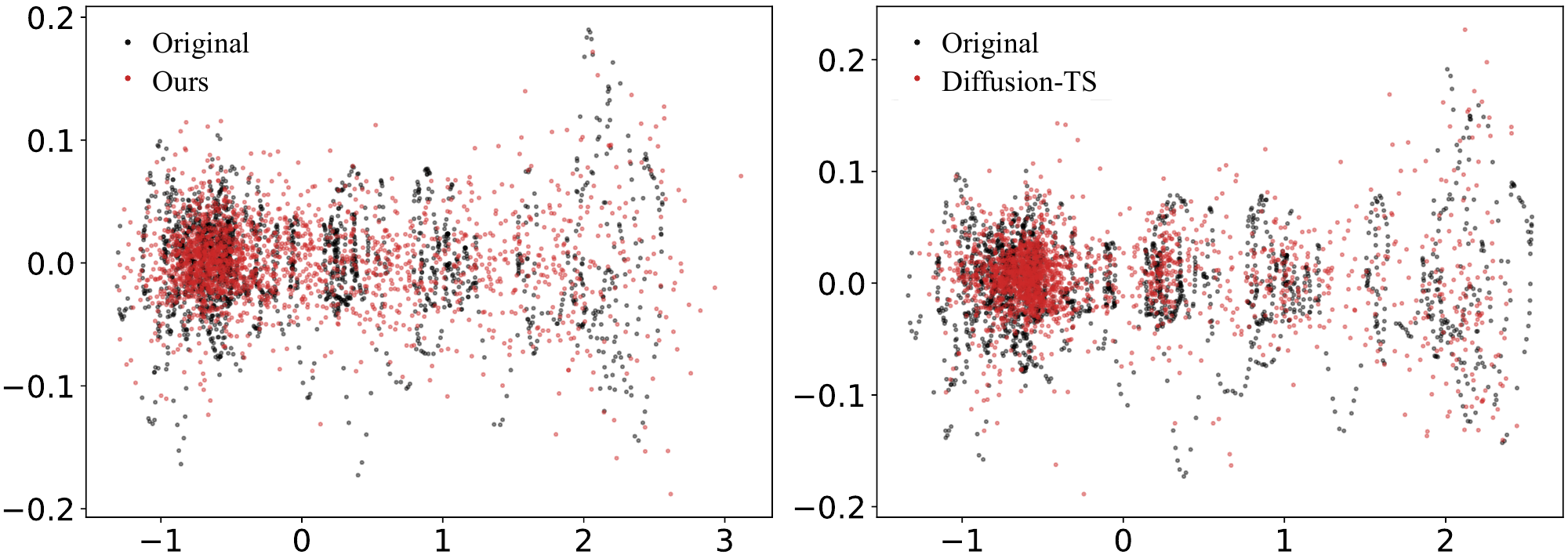}
    \caption{PCA analysis on Stocks dataset}
    \label{tsne_kde}
\end{wrapfigure}

To evaluate the quality of synthetic data, we conduct PCA, KDE, and t-SNE analyzes. As shown in Figure~\ref{tsne_kde}, PCA visualizations provide insights into the alignment between generated samples and real data in the reduced feature space. TimeFlow demonstrates a significantly closer clustering with the original distribution compared to Diffusion-TS, suggesting its stronger ability to retain the global variance structure of time series. The KDE plots in Figure~\ref{tsne_kde} further confirm this finding, showing that the marginal distributions of sequences produced by TimeFlow closely match the ground truth. These results demonstrate that TimeFlow generates more realistic and distributionally aligned time series, effectively capturing both global and local dynamics. Additional t-SNE visualizations are provided in the Appendix~\ref{visualization}.
 

\subsection{Condition Time Series Generation}

\begin{figure}[h]
    \centering
    \includegraphics[width=\textwidth]{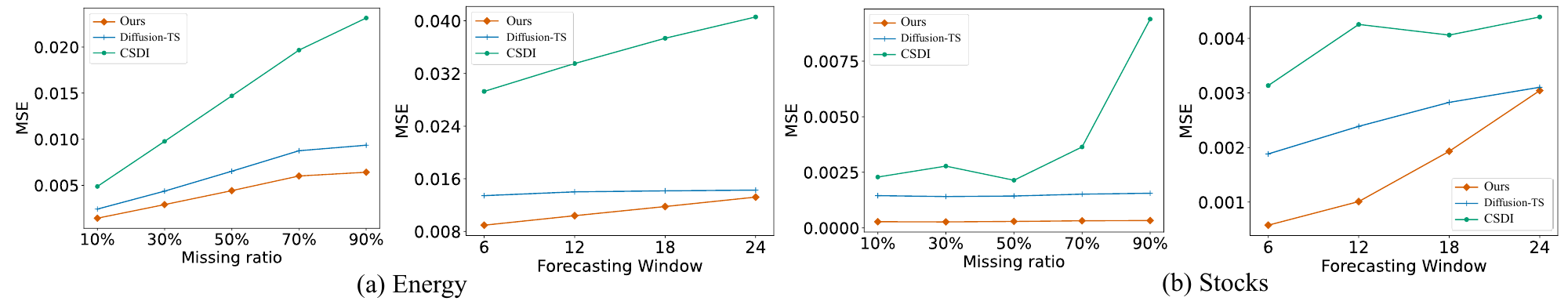}
    \caption{Performance of various methods for time-series imputation and forecasting.}
    \label{condition_figure1}
\end{figure}

We conduct extensive experiments to evaluate our model on both forecasting and imputation tasks. Figure~\ref{condition_figure1} presents results on the Energy and Stocks datasets with sequence length 48. For the imputation task, as the missing ratio increases from 0.1 to 0.9, TimeFlow consistently achieves lower MSE than the baselines, indicating its robustness in handling stochastic perturbations under severe data sparsity. For the forecasting task, as the prediction window expands from 6 to 24, our model maintains lower MSE than Diffusion-TS and CSDI, demonstrating its effectiveness in capturing intrinsic randomness in temporal dynamics while preserving long range predictive accuracy.

\begin{figure}[htp]
    \centering
    \includegraphics[width=\textwidth]{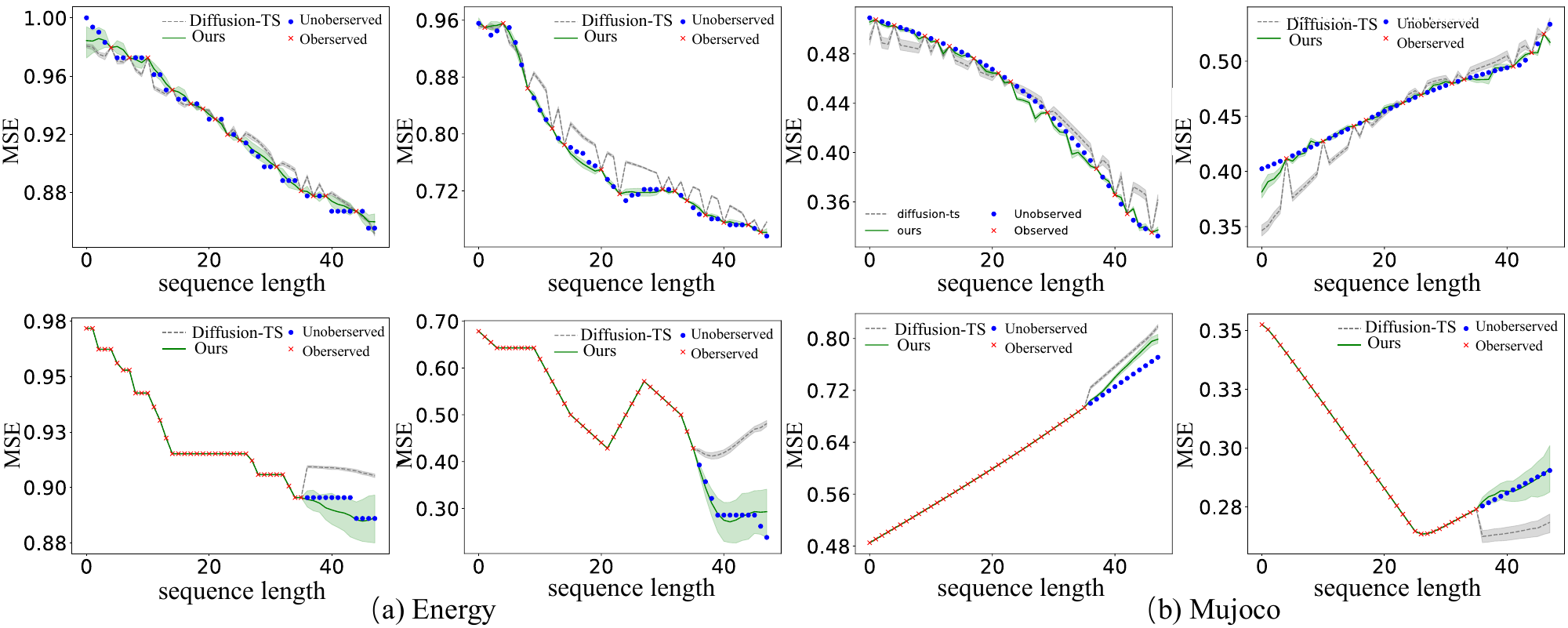}
    \caption{Examples of time series imputation and forcasting for Energy and Mujoco datasets.}
    \label{condition_figure2}
\end{figure}

To further assess the reliability of our method, we provide qualitative visualizations on the Energy and Mujoco datasets in Figure~\ref{condition_figure2}. The solid line denotes the median trajectory, while the shaded region indicates the 5\%–95\% quantile interval, capturing the uncertainty of the generated sequences. In the imputation task with a missing ratio of 0.9, our method successfully reconstructs coherent temporal patterns and smoothly recovers missing segments under extremely sparse observations. 
In the forecasting task with sequence length 48 and prediction length 12, our model produces trajectories more consistent with the ground truth and with fewer deviations than Diffusion-TS.

\vspace{-5pt}
\subsection{Effect of Diffusion Coefficient}
\begin{figure}[h]
    \centering
    \includegraphics[width=\textwidth]{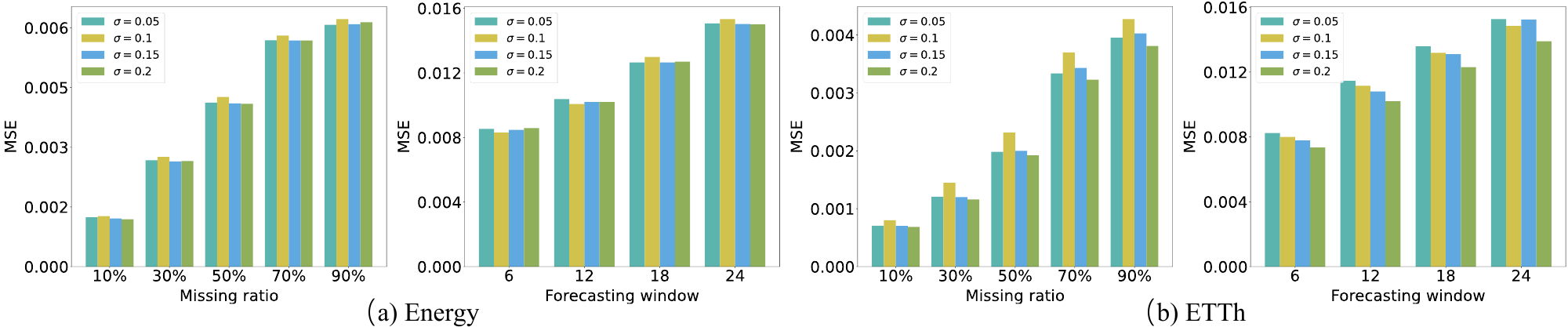}
    \caption{Impact of diffusion coefficient on conditional generation performance}
    \label{noise}
\end{figure}

We further investigate the impact of the constant diffusion coefficient in conditional generation. 
As shown in Figure~\ref{noise}, results on the Energy and ETTh datasets reveal distinct behaviors. 
Energy is a high-dimensional dataset with stable dynamics, where the influence of diffusion remains marginal and performance is largely consistent across noise levels. 
In contrast, ETTh has lower dimensionality with stronger variability, making the choice of diffusion coefficient critical. Moderate noise yields more reliable results by balancing uncertainty modeling and trajectory stability, whereas very small or large noise leads to underfitting or instability. These findings highlight that the effectiveness of noise injection depends on dataset characteristics, underscoring the importance of adaptive uncertainty modeling for robust conditional generation.

\vspace{-13pt}
\subsection{Analysis of Efficiency} 
\begin{figure}[htbp]
    \centering
    \begin{minipage}[b]{0.48\textwidth}
        \centering
        \includegraphics[width=\linewidth]{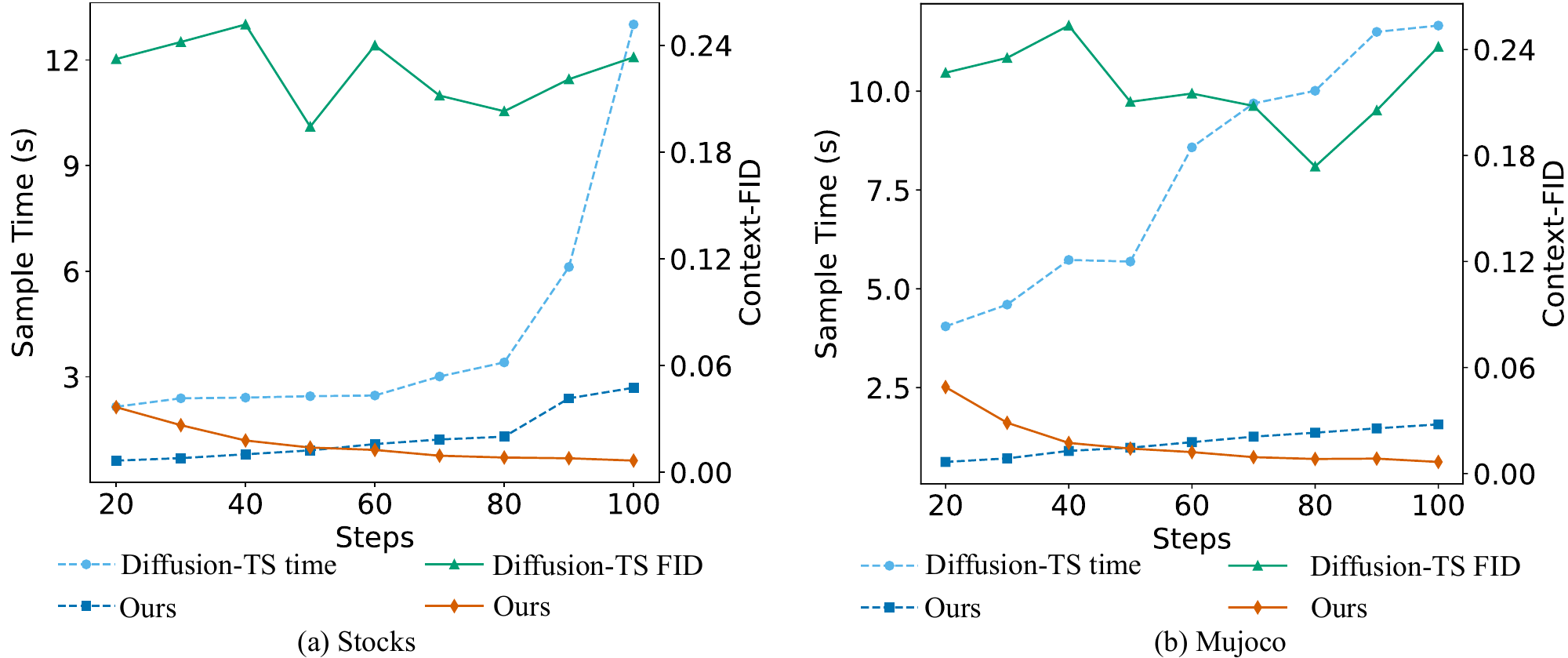}
        \captionof{figure}{Analysis of efficiency and generation metrics on Stocks dataset}
        \label{fig:efficiency}
    \end{minipage}
    \hfill
    \begin{minipage}[b]{0.48\textwidth}
        \centering
        \scriptsize
        \resizebox{\linewidth}{!}{%
        \begin{tabular}{c|c|c|c|c|c}
            \toprule
            Methods & Stage & Stocks & MuJoCo & Energy & fMRI \\
            \midrule
            \multirow{4}{*}{TimeFlow} 
                & train       & 373.88 & 600.18 & 1112.6 & 453.27 \\
                & uncondition & 2.69   & 5.16   & 25.52  & 5.69   \\
                & forecasting & 2.54 & 4.86 & 9.14 & 1.56 \\
                & imputation  & 2.08 & 4.73 & 9.22 & 3.62 \\
            \midrule
            \multirow{4}{*}{Diffusion-ts} 
                & train       & 687.69 & 873.39 & 1690.77 & 983.48 \\
                & uncondition & 3.08   & 7.65   & 28.11   & 17.69  \\
                & forecasting & 29.41 & 39 & 123.17 & 71.84 \\
                & imputation  & 29.99 & 41.36 & 101.91 & 122.02 \\
            \bottomrule
        \end{tabular}}
        \vspace{2pt}
        \captionof{table}{Training and sampling time comparison on different datasets in seconds (s).}
        \label{tab:runtime}
    \end{minipage}
\end{figure}

\vspace{-5pt}
To validate the efficiency of our model, we comprehensively compare TimeFlow with Diffusion-TS across different sampling steps and diverse datasets. On Stocks and Mujoco, TimeFlow requires significantly less sampling time and achieves superior Context-FID under the same step size (Figure~\ref{fig:efficiency}). Moreover, across multiple datasets, TimeFlow consistently demonstrates clear efficiency advantages, with lower training costs and faster sampling than Diffusion-TS (Table~\ref{tab:runtime}).
\vspace{-5pt}

\vspace{-15pt}
\subsection{Ablation}

To evaluate the effectiveness of the proposed TimeFlow model, we conduct an ablation study by removing three key components: (1) w/o CA: removing cross-attention in the flow decomposition; (2) w/o FD: removing the decomposition mechanism; and (3) w/o Encoder: removing the self-attention backbone. The results are summarized in Table~\ref{ablation}.We observe that the complete TimeFlow consistently delivers superior or competitive performance across all datasets and evaluation metrics. Removing the CA or FD leads to only a moderate degradation in performance. Interestingly, on high-dimensional datasets such as fMRI, eliminating the encoder unexpectedly improves performance. We attribute this to the fact that, in extremely high-dimensional scenarios, the encoder may introduce excessive complexity and amplify noise, which can hinder generalization.
\vspace{-5pt}
\begin{table}[htbp]
  \centering
  \scriptsize
  \caption{Ablation study for model architecture and options. (Bold indicates best performance).}
  \begin{tabular}{c|c|c|c|c|c|c|c}
    \toprule
    Metric & Methods & Sines & Stocks & ETTh & MuJoCo & Energy & fMRI \\
    \midrule
    \multirow{4}{*}{\makecell{Discriminative \\ Score}} 
        & TimeFlow      & \textbf{0.004±0.004} & \textbf{0.011±0.001} & \textbf{0.010±0.006} & \textbf{0.005±0.007} & \textbf{0.061±0.007} & 0.101±0.015 \\
        & w/o CA    & 0.005±0.004 & 0.017±0.013 & 0.014±0.012 & 0.010±0.010 & 0.063±0.014 & 0.094±0.016 \\
        & w/o FD  & 0.005±0.005 & 0.092±0.029 & 0.013±0.005 & 0.012±0.019 & 0.062±0.013 & 0.149±0.030 \\
        & w/o Encoder   & 0.010±0.004 & 0.104±0.009 & 0.012±0.007 & 0.101±0.077 & 0.142±0.052 & \textbf{0.067±0.098} \\
    \midrule
    \multirow{4}{*}{\makecell{Predictive \\ Score}} 
        & TimeFlow      & \textbf{0.093±0.000} & \textbf{0.037±0.000} & \textbf{0.119±0.002} & \textbf{0.008±0.000} & \textbf{0.250±0.000} & \textbf{0.099±0.000} \\
        & w/o CA    & \textbf{0.093±0.000} & \textbf{0.037±0.000} & 0.121±0.004 & 0.010±0.000 & 0.251±0.000 & 0.100±0.000 \\
        & w/o FD  & \textbf{0.093±0.000} & 0.038±0.000 & 0.123±0.003 & 0.011±0.000 & 0.251±0.000 & 0.100±0.000 \\
        & w/o Encoder    & 0.093±0.001 & 0.039±0.000 & 0.121±0.005 & 0.015±0.002 & 0.252±0.000 & 0.100±0.000 \\
    \bottomrule
  \end{tabular}
  \label{ablation}
\end{table}
\vspace{-10pt}

\vspace{-10pt}
\section{Conlusion}
\vspace{-5pt}
In this paper, we propose TimeFlow, a novel flow matching framework under the SDE paradigm for time series generation. By introducing a constant diffusion coefficient, TimeFlow effectively captures stochasticity and improves robustness. The flow matching formulation enables efficient training and sampling while maintaining high-quality generation. Experiments on multiple datasets demonstrate its effectiveness in both unconditional and conditional settings. Future work will explore more efficient solvers and scaling TimeFlow to large-scale applications.




\newpage
\bibliography{iclr2026_conference}
\bibliographystyle{iclr2026_conference}
\newpage


\end{document}